\documentclass[runningheads]{llncs}
\usepackage{graphicx}
\usepackage[utf8]{inputenc}
\usepackage{epsfig}
\usepackage{calc}
\usepackage{amssymb}
\usepackage{amstext}
\usepackage{amsmath}
\usepackage{amsfonts}
\usepackage{multicol}
\usepackage{pslatex}
\usepackage{listings}
\usepackage[ruled,vlined,linesnumbered]{algorithm2e}
\usepackage{comment}
\usepackage{url}
\usepackage{enumitem}   
\usepackage{times}
\usepackage{latexsym}
\usepackage{graphicx}
\usepackage{color}
\usepackage{multirow}

\begin{document}

\title{Heuristically Guided Compilation for Multi-Agent Path Finding}
\titlerunning{Heuristically Guided Compilation for Multi-Agent Path Finding}


%
%
\author{Pavel Surynek\orcidID{0000-0001-7200-0542}}
%
\authorrunning{P. Surynek}

%
\institute{
Faculty of Information Technology\\
Czech Technical University in Prague\\
Th\'{a}kurova 9, 160 00 Praha 6, Czechia\\
\email{pavel.surynek@fit.cvut.cz}
}


\maketitle

\begin{abstract}
Multi-agent path finding (MAPF) is a task of finding non-conflicting paths connecting agents' specified initial and goal positions in a shared environment. We focus on compilation-based solvers in which the MAPF problem is expressed in a different well established formalism such as mixed-integer linear programming (MILP),  Boolean satisfiability (SAT), or constraint programming (CP). As the target solvers for these formalisms act as black-boxes it is challenging to integrate MAPF specific heuristics in the MAPF compilation-based solvers. We show in this work how the build a MAPF encoding for the target SAT solver in which domain specific heuristic knowledge is reflected. The heuristic knowledge is transferred to the SAT solver by selecting candidate paths for each agent and by constructing the encoding only for these candidate paths instead of constructing the encoding for all possible paths for an agent. The conducted experiments show that heuristically guided compilation outperforms the vanilla variants of the SAT-based MAPF solver.

\end{abstract}

\section{Introduction}

Multi-agent path finding (MAPF) represents a fundamental problem in artificial intelligence \cite{DBLP:conf/aiide/Silver05,DBLP:conf/ijcai/Ryan07,DBLP:conf/aaai/Standley10,DBLP:conf/ijcai/LunaB11,DBLP:conf/icra/YuL13}. The task is to navigate each agent from the set of agents $A=\{a_1,a_2,...,a_k\}$ from its initial position to a specified goal position. In the standard discrete variant of MAPF, the environment is modeled as an undirected graph $G=(V,E)$ where vertices represent positions and edges represent the topology across which the agents move between vertices. There are two requirements that make the MAPF problem challenging: {\bf (1)} the agents must not collide with each other, that is they never can share a vertex nor can traverse an edge in opposite directions and {\bf (2)} an objective such as the total number of move actions must be optimized.

There are many real-life applications of MAPF ranging from motion planning in robotics where agents are represented by actual robots to computer games where agents are represented by virtual characters. Other examples include discrete multi-agent navigation and coordination \cite{DBLP:conf/iros/LunaB10}, agent rearrangement in automated warehouses \cite{DBLP:journals/tase/BasileCC12}, ship collision avoidance \cite{DBLP:journals/jaciii/KimHP14}, or formation maintenance and maneuvering of aerial vehicles \cite{DBLP:conf/icra/ZhouS15}.

We address the MAPF problem from the perspective of compilation-based techniques. {\em Compilation} is one of the most important techniques used across computing. In the context of problem solving in artificial intelligence, the compilation approach reduces an input problem instance from its source formalism to a different, usually well established, target formalism for which an efficient solver exists. After obtaining a solution by the solver in the target formalism, the solution is interpreted back to the input formalism, which altogether constitutes a {\bf reduction-solving-interpretation} loop.

Target formalisms are often combinatorial optimization frameworks like {\em constraint programming / optimization} (CP) \cite{DBLP:books/daglib/0016622}, {\em mixed integer linnear programming} (MILP) \cite{DBLP:books/daglib/0023873,rader2010deterministic}, Boolean satisfiability (SAT) \cite{DBLP:series/faia/2009-185}, satisfiability modulo theories (SMT) \cite{DBLP:reference/mc/BarrettT18}, or answer set programming (ASP) \cite{DBLP:books/sp/Lifschitz19}. Taking advantage of the progress in solvers for the target formalisms, often accumulated for decades, in solving the input problem represents the key benefit of problem solving via compilation. However, the way how the input instance is reduced to the target formalism and presented to the solver has a great impact on the efficiency of the whole solving process.

Currently {\bf compilation-based} optimal solvers for MAPF represent a major alternative to {\bf search-based} solvers, that model and solve the problem directly, and often provide more modular and versatile architecture than search-based solvers while providing competitive performance. Contemporary compilation-based solvers for MAPF include those based on CP \cite{DBLP:conf/aips/GangeHS19}, MILP \cite{DBLP:conf/ijcai/LamBHS19}, SAT/SMT \cite{DBLP:conf/socs/SurynekFSB16,DBLP:conf/ijcai/Surynek19}, as well as ASP-based solvers \cite{DBLP:conf/aaai/ErdemKOS13}.

We focus in this paper on a recent SMT-based approach to solving MAPF \cite{DBLP:conf/iros/Surynek21} that uses multi-level lazy compilation of MAPF to Boolean satisfiability. The previous approach builds on top the SMT-CBS, an optimal algorithm, \cite{DBLP:conf/ijcai/Surynek19}, that encodes an incomplete specification of the input MAPF instance into SAT. Concretely, collision avoidance constraints are not encoded at the beginning which may result in solutions that lead to a collisions. After the collisions are detected, collision avoidance constraints are added to the encoding and the process is repeated until the collision-free solution is obtained. The advantage of the lazy approach is that a collision-free solution is often obtained well before all collision avoidance constraints are encoded.

The further step from SMT-CBS is the next level of laziness called {\em sparsification} that adds laziness in encoding of the set of candidate paths for each agent. While the original SMT-CBS gives each agent a chance to chose any of the possible paths, the sparse version starts with a restricted set of candidate paths containing only the most promising path. If the search for non-colliding paths with sparse set of candidate paths is unsuccessful, the set of candidate paths is extended. The process may end up with all paths included but again the solution is often found before all paths are considered.

Since Boolean formulae, to which the input MAPF instance is reduced in SMT-CBS, are derived from the set of candidate paths, the effect of sparsification of the set is twofold: {\bf (1)} it leads to smaller target Boolean formulae that can be constructed faster and {\bf (2)} the satisfiability of formulae can be decided by the SAT solver faster, altogether improving the reduction-solving-interpretation loop in SMT-CBS.
 
The original sparsification technique used the selection of candidate paths so that at least one path for any subset of detected collisions is taken. Since the collisions are treated using the `or' connective in this approach,  we call it an {\em OR-path selection heuristic}.

 In this work, we contribute by a different heuristically guided selection of candidate paths. I contrast to OR-path selection, we suggest to include one path that avoids all recenlty discovered collisions. As collisions are threated via the `and' connective, we call our new approach an {\em AND-path selection heuristic}.
 
 Since candidate paths are represented by {\em multi-value decision diagram} (MDD), intoducing new vertices of the newly represented path results in representing many more other paths giving the agent more chances to avoid the collision. On the other hand, compared to the OR-path selection, AND-path selection adds fewer new vertices to MDD and thus sparsification via AND-path seleciton is more aggresive. We conjecture that AND-path selection will yield smaller MDDs and hence smaller formulae in the course of execution of SMT-CBS.

The paper is organized as follows: the MAPF problem and basic compilation-based methods for MAPF are introduced first. Then the concept of lazy compilation schemes is described from which the heuristically guided sparsification, the main contribution, is derived. The new algorithm based on the new heuristic, called Heuristic-SMT-CBS, is developed next and the experimental evaluation follows.

\section{Background}

In this section we introduce the problem of path planning for multiple agents formally and present optimal solving algorithms with focus on compilation-based techniques.

\subsection{Multi-agent Path Finding}

We assume discrete time in MAPF. The configuration of agents at timestep $t$ is denoted as $s_t$. Each agent $a_i$ has a start position $s_0(a_i) \in V$ and a goal position $s_+(a_i) \in V$. 

\begin{figure}[h]
    \centering
    \includegraphics[trim={3.2cm 22cm 4.8cm 2.3cm},clip,width=0.9\textwidth]{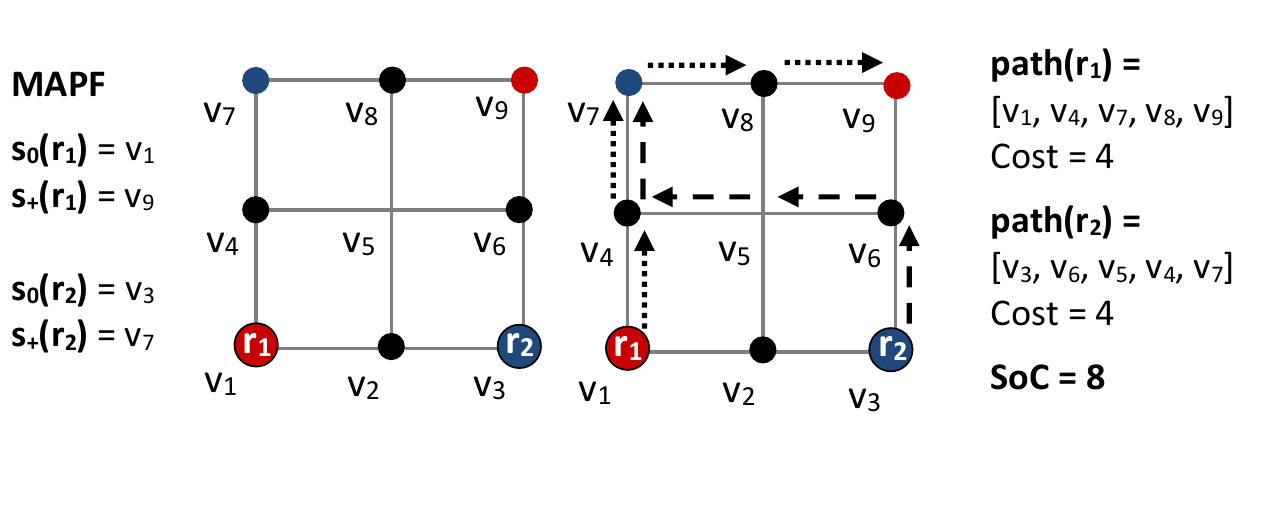}
    \vspace{-1.0cm}\caption{An MAPF instance with two agents $a_1$ and $a_2$.}
    \label{figure-MAPF}
\end{figure}

At each time step an agent can either {\em move} to an adjacent vertex or {\em wait} in its current vertex. The task is to find a sequence of move/wait actions for each agent $a_i$, moving it from $s_0(a_i)$ to $s_+(a_i)$ such that agents do not {\em collide}, i.e., do not occupy the same location at the same time and do not traverse the same edge in opposite directions. An example of MAPF instance and its solution is shown in Figure \ref{figure-MAPF}.

Formally, a MAPF instance is a tuple $\Sigma=(G=(V,E),R,s_0,s_+)$ where $s_0:R \rightarrow V$ is an initial configuration of agents and $s_+:R \rightarrow V$ is a goal configuration of agents. A \textit{solution} for $\Sigma$ is a sequence of configurations $\mathcal{S}(\Sigma)=[s_0,s_1,...,s_{\mu}]$ such that $s_{t+1}$ results from valid movements from $s_{t}$ for $t=1,2,...,\mu-1$, and $s_{\mu}=s_+$. Orthogonally to this, the solution can be represented as a set of paths for individual agents. 

Various cumulative objectives are often optimized in MAPF. We will develop all concepts in this paper for the {\em sum-of-costs} objective,  denoted $\mathit{SoC}$. $\mathit{SoC}$ is the summation, over all agents, of the number of time steps required to reach the goal. Formally, $\mathit{SoC} = \sum_{i=1}^k{\mathit{cost}(path(a_i))}$, where $\mathit{cost}(path(a_i))$ is an \textit{individual path cost} of agent $a_i$ connecting $s_0(a_i)$ calculated as the number of edge traversals and wait actions. \footnote{The notation $path(a_i)$ refers to path in the form of a sequence of vertices and edges connecting $s_0(a_i)$ and $s_+(a_i)$ while $\mathit{cost}$ assigns the cost to a given path.}

The sum-of-costs objective accumulates the cost of wait actions for agents not yet reaching their goal vertices. A feasible solution of a solvable MAPF instance can be found in polynomial time \cite{DBLP:conf/focs/KornhauserMS84}.

Finding an optimal solution with respect to the sum-of-costs objective is NP-hard \cite{DBLP:conf/aaai/YuL13,DBLP:conf/aaai/Surynek10} and also determining the existence of a solution that differs from the optimum by a factor less than $4/3$ is NP-hard too \cite{DBLP:conf/aaai/MaTSKK16}. Therefore designing algorithms based on search and SAT for MAPF is justifiable.

\subsection{Compilation-based Approaches}

The idea behind compilation that uses the SAT paradigm is to construct a Boolean formula whose satisfiability corresponds to existence of a solution of the given value of the objective, that is, the formula encodes bounded instance of MAPF. In our case we are encoding the question whether there is a solution of sum-of-costs $\mathit{SoC}$ to a given MAPF $\Sigma$. Moreover, the approach is constructive; that is, the formula exactly reflects the MAPF instance and if it is satisfiable, solution of MAPF can be reconstructed from satisfying assignment of the formula.

There are two ways how to connect satisfiability of the formula and solvability of $\Sigma$: using either {\bf equivalence} or {\bf implication}.

We say  $\mathcal{F(\mathit{SoC})}$ to be a {\em complete Boolean model} of MAPF.

\begin{definition}
  {\bf (complete Boolean model)} Boolean formula $\mathcal{F(\mathit{SoC})}$ is a {\em complete Boolean model} of MAPF $\Sigma$
  if the following condition holds: $\mathcal{F(\mathit{SoC})}$ is satisfiable $\Leftrightarrow$ $\Sigma$ has a solution of sum-of-costs $\mathit{SoC}$.
\end{definition}

Complete Boolean models were the used in makespan optimal SAT-based solvers for MAPF \cite{DBLP:journals/amai/Surynek17} and in MDD-SAT \cite{DBLP:conf/ecai/SurynekFSB16}, the first sum-of-costs optimal SAT-based solver. A natural relaxation from the complete Boolean model is an {\em incomplete Boolean model} where instead of the equivalence between solving MAPF and the formula we require an implication only. Incomplete models are inspired from the SMT paradigm and are used in the recent sum-of-costs optimal solver SMT-CBS \cite{DBLP:conf/ijcai/Surynek19,DBLP:conf/iros/Surynek21}.

\begin{definition}
  {\bf (incomplete Boolean model).} Boolean formula $\mathcal{H(\mathit{SoC})}$ is an {\em incomplete Boolean model} of MAPF 
  $\Sigma$ if the following condition holds: $\mathcal{H(\mathit{SoC})}$ is satisfiable $\Leftarrow$ $\Sigma$ has a solution of sum-of-costs $\mathit{SoC}$.
\end{definition}

Being able to construct formula $\mathcal{F}$ one can obtain optimal MAPF solution by checking satisfiability of $\mathcal{F}(0)$, $\mathcal{F}(1)$, $\mathcal{F}(2)$,... until the first satisfiable $\mathcal{F(\mathit{SoC})}$ is met. This is possible due to monotonicity of MAPF solvability with respect to increasing values of common cumulative objectives such as the sum-of-costs. In practice it is however impractical to start at 0; lower bound estimation is used instead - sum of lengths of shortest paths can be used in the case of sum-of-costs.

Construction of $\mathcal{F(\mathit{SoC})}$ relies on the time expansion of underlying graph $G$. Having $\mathit{SoC}$, the basic variant of time expansion determines the maximum number of time steps $\mu$ (also refered to as a {\em makespan}) such that every possible solution of the given MAPF with the sum-of-costs less than or equal to $\mathit{SoC}$ fits within $\mu$ timesteps (that is, no agent is outside its goal vertex after $\mu$-th timestep if the sum-of-costs $\mathit{SoC}$ is not to be exceeded).

The time expansion itself makes copies of vertices $V$ for each timestep $t=0,1,2,...,\mu$. That is, we have vertices $v^t$ for each $v \in V$ time step $t$. Edges from $G$ are converted to directed edges interconnecting timesteps in time expansion. Directed edges $(u^t,v^{t+1})$ are introduced for $t=1,2,...,\mu-1$ whenever there is $\{u,v\} \in E$. Wait actions are modeled by introducing edges $(u^t,t^{t+1})$. A directed path in time expansion corresponds to trajectory of a agent in time. Hence the modeling task now consists in construction of a formula in which satisfying assignments correspond to directed paths from $s_0^0(a_i)$ to $s_+^\mu(a_i)$ in the time expansion.

Assume that we have time expansion $(V_i,E_i)$ for agent $a_i$. Propositional variable $\mathcal{X}_v^t(a_j)$ is introduced for every vertex $v^t$ in $V_i$. The semantics of $\mathcal{X}_v^t(a_i)$ is that it is $\mathit{TRUE}$ if and only if agent $a_i$ resides in $v$ at time step $t$. Similarly we introduce $\mathcal{E}_{u,v}^t(a_i)$ for every directed edge $(u^t,v^{t+1})$ in $E_i$. Analogically the meaning of $\mathcal{E}_{u,v}^t(a_i)$ is that it is $\mathit{TRUE}$ if and only if agent $a_i$ traverses edge $\{u,v\}$ between time steps $t$ and $t+1$.

Once we have the Boolean decision variables $\mathcal{X}_v^t(a_i)$ and $\mathcal{E}_{u,v}^t(a_i)$ we can introduce constraints so that truth value assignments are restricted only to those that correspond to valid solutions of a given MAPF. The added constraints together ensure that $\mathcal{F(\mathit{SoC})}$ is a {\em complete propositional model} for given MAPF.

We here illustrate the model by showing few representative constraints. For the detailed list of constraints we refer the reader to \cite{DBLP:conf/ecai/SurynekFSB16}.

Collisions between agents can be eliminated by the following constraint over $\mathcal{X}_v^t(a_i)$ variables for every $v \in V$ and timestep $t$:

\begin{equation}
    {\sum_{a_i \in A \:|\:v^t \in V_i}{\mathcal{X}^t_v(a_i)} \leq 1
    }
    \label{eq-1}
\end{equation}

Next, there is a constraint stating that if agent $a_i$ appears in vertex $u$ at time step $t$ then it has to leave through exactly one edge $(u^t,v^{t+1})$. This can be established by following constraints:

\begin{equation}
   {  \mathcal{X}_u^t(a_i) \Rightarrow \bigvee_{(u^t,v^{t+1}) \in E_i}{\mathcal{E}^t_{u,v}(a_i),}
   }
   \label{eq:basic-start}
\end{equation}
\begin{equation}
   {  \sum_{v^{t+1}\:|\:(u^t,v^{t+1}) \in E_i }{\mathcal{E}_{u,v}^t{(a_i)} \leq 1}
   }
   \label{eq-2}
\end{equation}

Other constraints ensure that truth assignments to variables per individual agents form paths. That is if agent $a_i$ enters an edge it must leave the edge at the next time step.

\begin{equation}
   {  \mathcal{E}^t_{u,v}(a_i) \Rightarrow \mathcal{X}^t_v(a_i) \wedge \mathcal{X}^{t+1}_v(a_i)
   }
   \label{eq-4}
\end{equation}

A common measure how to reduce the number of decision variables derived from the time expansion is the use of {\em multi-valued decision diagrams} (MDDs) \cite{DBLP:conf/cp/AndersenHHT07,DBLP:journals/ai/SharonSGF13}. The basic observation that holds for MAPF is that an agent can reach vertices in the distance $d$ (distance of a vertex is measured as the length of the shortest path) from the current position of the agent no earlier than in the $d$-th time step. Analogical observation can be made with respect to the distance from the goal position.

\begin{figure}[h]
    \centering
    \vspace{-0.2cm}
    \includegraphics[trim={2.7cm 16.2cm 2.8cm 3cm},clip,width=0.9\textwidth]{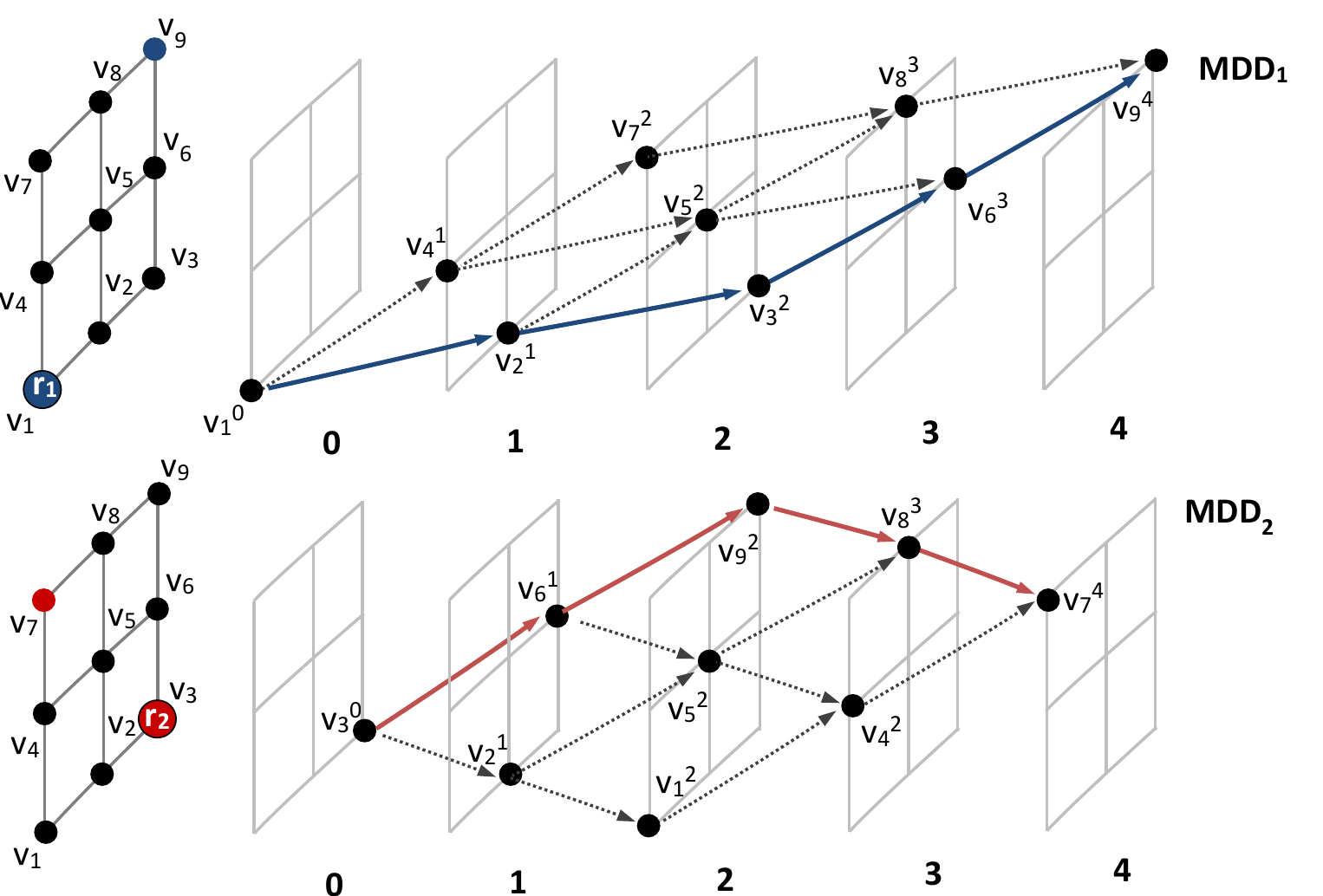}
    \vspace{-0.2cm}
    \caption{An example of MDDs representing all candidate paths of cost 4 for agents $a_1$ nad $a_2$ \cite{DBLP:conf/iros/Surynek21}.}
    \label{fig-MDD}
\end{figure}

Above observations can be utilized when making the time expansion of $G$. For a given agent, we do not need to consider all vertices at time step $t$ but only those that are reachable in $t$  timesteps from the initial position and that ensure that the goal can be reached in the remaining $\mu - t$ timesteps. Time expansion using MDDs is shown in Figure \ref{fig-MDD}.

The combination of SAT-based approach and MDD time expansion is the basis for the MDD-SAT algorithm. It is important to note that from out perspective the MDD can be regarded as a representation of the set of candidate paths for given agent. In MDD-SAT, MDDs represent all possible paths that meet the given makespan and sum-of-costs bounds.

\subsection{CBS and Lazy Compilation}

{\em Conflict-based search} (CBS) \cite{DBLP:journals/ai/SharonSFS15} is a popular algorithm for solving MAPF optimally. From the compilation perspective, the CBS algorithm could be understood as a {\bf lazy} method that tries to solve an underspecified problem and relies on to be lucky to find a correct solution even using this incomplete specification. There is another mechanism that ensures soundness of this lazy approach, the branching scheme. If the CBS algorithm is not lucky, that is, the candidate solution is incorrect in terms of the MAPF rules, then the search branches for each possible refinement of discovered MAPF rule violation and the refinement is added to the problem specification in each branch. Concretely, the MAPF rule violations are conflicts of pairs of agents such as collision of $a_i \in A$ and $a_j \in A$ in $v$ at time step $t$ and the refinements are conflict avoidance constraints for single agents in the form that $a_i \in A$ should avoid $v$ at time step $t$ (for $a_j$ analogously).

Analogously the avoidance for edge conflicts can be done. For the sake of brevity, all the concepts developed further will be introduced only for vertex conflicts.

The idea of laziness of CBS has been combined with compilation-based approaches in SMT-CBS algorithm. The major difference from the standard CBS is that there is no branching at the high level. The high level SMT-CBS roughly correspond to the main loop of MDD-SAT, that is, the algorithm checks the existence of MAPF solution for the sum of costs $\mathit{SoC}_0$, $\mathit{SoC}_0+1$, $\mathit{SoC}_0 +2$, ..., where $\mathit{SoC}_0$ is the sum of lengths of shortest paths connecting agents' start and goal positions, until the algorithm gets a positive answer from the SAT solver. SMT-CBS differs from MDD-SAT in using the incomplete Boolean model in which it ignores specific between agents at beginning. Collisions are resolved lazily similarly as in CBS.

While the collision between $a_i$ and $a_j$ in vertex $v$ at time step $t$ in standard CBS is resolved as high-level branching is, in SMT-CBS it is resolved by refinement of $\mathcal{F}(\mathit{SoC})$ with a new disjunction $\neg \mathcal{X}_v^t(a_i) \vee \neg \mathcal{X}_v^t(a_j)$. Branching is thus deferred into the SAT solver. The advantage of SMT-CBS is that it builds the formula lazily; that is, it adds constraints on demand after a conflict occurs. Such approach may save resources as solution may be found before all constraint are added.


\section{Heuristically Guided Sparsification in Lazy Compilation}

Although lazy compilation in SMT-CBS is an important factor that reduces the size of generated Boolean formulae in SMT-CBS compared to MDD-SAT, the reduction concerns only the conflict avoidance constraints for pairs of agents (binary disjunctions). The number of nodes as well as the number of directed edges in the MDDs is not reduced by laziness and hence the number of variables and constraints modeling the existence of directed paths in MDDs in the target Boolean formula is the same as in MDD-SAT. For instances that take place on large graphs, the size of MDDs directly reflected in the size of the formula could still be prohibitive.

It has been suggested in \cite{DBLP:conf/iros/Surynek21} as part of the Sparse-SMT-CBS algorithm to reduce the number of directed paths being encoded. This is done by using sparse sets of candidate paths for each agent that satisfy given sum-of-costs and makespan bounds. That is, instead of considering all such paths as done in MDDs only a relevant subset of them is taken into account. The sparse set of candidate paths is denoted $\Pi$ or $\Pi(a_i)$ for specific agent $a_i$. Similar concept called the {\em pool of paths} has been used in the context of MILP-based compilation for MAPF \cite{DBLP:conf/aips/GangeHS19} which however does not explicitly focus on sparsification.

Sparse-SMT-CBS uses MDDs to represent the set of candidate paths, to distinguish it from MDD containing all paths it is refered to as Sparse MDD or SMDD in short. Given the maximum number of steps $\mu$ and any (sparse) set of paths $\Pi(a_i)$ in the time expansion of $G$ of length at most $\mu$, we are able to construct SMDD$_i$ that represents $\Pi(a_i)$. That is, any path from $\Pi(a_i)$ is represented in SMDD$_i$. This is done by including corresponding nodes into SMDD and interconnecting them by directed edges.

It is however important to note that SMDD tends to overestimate the set of paths being represented, that is, the super-set of $\Pi(a_i)$ is actually represented as shown in the following example:

\begin{example}
\label{ex:smdd}
Consider $\Pi(a_i)$ consisting of two paths $[v_1,v_2,v_3,v_4,v_5]$ and $[v_1,v_6,v_3,v_7,v_5]$. Once there paths are represented in SMDD, this yields nodes $v_1^0,v_2^1,v_3^2,v_4^3,v_5^4,v_6^1$ and $v_7^3$ in SMDD and directed edges: $(v_1^0,v_2^1);(v_2^1,v_3^2);(v_3^2,v_4^3);(v_4^3,v_5^4);(v_1^0,v_6^1);(v_6^1,v_3^2);(v_3^2,v_7^3)$ and $(v_7^3,v_5^4)$. Despite only two paths we originally represented, the resulting SMDD represent four paths: $[v_1,v_2,v_3,v_4,v_5]$, $[v_1,v_6,v_3,v_4,v_5]$, $[v_1,v_2,v_3,v_7,v_5]$, and $[v_1,v_6,v_3,v_7,v_5]$.
\end{example}

Sparse-SMT-CBS uses identical sum-of-costs and makespan bounds increasing scheme as SMT-CBS at the high-level. Each iteration at the high-level resolves a question whether there exists a solution to the input MAPF $\Sigma$ such that it fits in the current sum-of-costs $\mathit{SoC}$ and makespan $\mu$. This question is compiled as a series of Boolean formulae and consulted with the SAT solver at the low-level.

The low-level in Sparse-SMT-CBS is different from SMT-CBS. In both algorithms it tries to find a non-conflicting set of paths satisfying $\mathit{SoC}$ and $\mu$, but the set of candidate paths from which SMT-CBS selects is fixed in advance in MDD, while Spare-SMT-CBS starts with a minimal set of candidate paths and each time a new conflict is discovered the set of candidate paths is extended to reflect the new conflict. Spare-SMT-CBS does so by the {\em OR-path} selection heuristic that finds a path for {\bf every subset} of the set of the of conflicts and represents this path in the next SMDD.

\begin{figure}[h]
    \centering
    \vspace{-0.2cm}
    \includegraphics[trim={2.5cm 16.4cm 1.6cm 2.6cm},clip,width=0.9\textwidth]{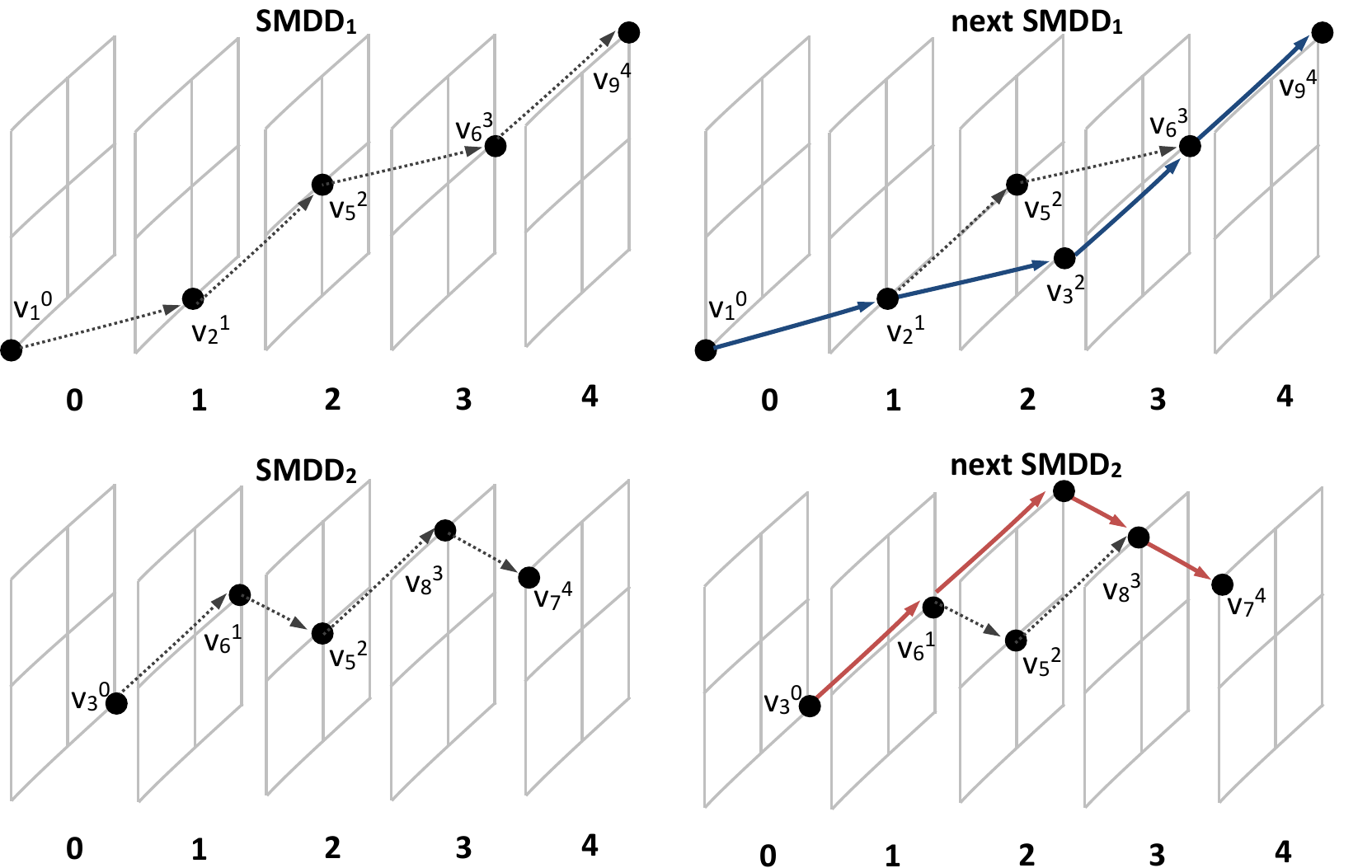}
    \vspace{-0.2cm}
    \caption{An example of sparse MDDs (SMDDs) for agents $a_1$ and $a_2$. The first iteration yields a conflict in $v_5$ at timestep 2 between the agents which can be avoided via newly represented paths in the next iteration.}
    \label{fig-sparseMDD}
\end{figure}

The example of SMDDs from the running example shown in Figure \ref{fig-sparseMDD}. Initially shortest paths are included in $\Pi(a_i)$ connecting agent's starting vertex and its goal, one shortest path per agent. However, the initial choice of paths is poor leading to a collision in $v_5$ at timestep 2 which is detected by Sparse-SMT-CBS and alternative paths reflecting the conflict are suggested for each agent by the new-Paths function. The new paths are included in SMDDs as shown in the right part of Figure \ref{fig-sparseMDD}. For these new SMDDs, non-conflicting path can be found.

Despite the extension of SMDD with paths suggested by the OR-path selection heuristic yields SMDD that is smaller than the full MDD. The SMDD can still be very large for the growing number of conflicts. This is due to the overestimating nature of SMDD as shown in Example \ref{ex:smdd}. Therefore we suggested a new heuristic called {\em AND-path} selection that adds to SMDD only one path that avoids all conflicts. Taking into account the overestimation of SMDD, even one path can add enough freedom of navigation for the agent using paths being represented by the SMDD. If such path does not exists then the algorithm switches to the full MDD mode that represents all paths satisfying given makespan and sum-of-costs bounds (which corresponds to previous SMT-CBS).


We integrated the SMDD reasoning and the AND-path selection heuristic into the SMT-CBS framework, designing a new algorithm we called {\bf Heuristic-SMT-CBS}. The pseudo-code of Heuristic-SMT-CBS is shown as Algorithm \ref{alg-heuristic-SMTCBS}. The encoding used in Heuristic-SMT-CBS follows MDD-SAT with collision avoidance constraints omitted but instead of deriving decision Boolean variables and constraints from MDDs they are derived from SMDDs. That is, a Boolean variable $\mathcal{X}(a_i)_v^t$ is introduced for each node $v^t$ in SMDD$_i$ and a variable $\mathcal{E}(a_i)_{u,v}^t$ is introduced for each edge in SMDD$_i$. Constraints are introduced analogously. Observe that the sparser the SMDDs are the smaller Boolean model is obtained.


\begin{algorithm}[!h]
\begin{footnotesize}
\SetKwBlock{NRICL}{Heuristic-SMT-CBS ($\Sigma = (G=(V,E),R,s_0,s_+))$}{end} \NRICL{
    $\mathit{conflicts} \gets \emptyset$\\
    $\Pi \gets$ $\{\Pi^*(a_i)$ a shortest path from $s_0(a_i)$ to $s_+(a_i) | i = 1,2,...,k\}$ \\
    $\mathit{SoC} \gets \sum_{i=1}^k{\mathit{cost}(\Pi(a_i))}$ \\
    $\mu \gets \max_{i=1}^k{\mathit{cost}(\Pi(a_i))}$ \\    
    \While {$\mathit{TRUE}$}{
         $(\mathit{path},\mathit{conflicts}) \gets$ Heuristic-SMT-CBS-Fixed($\mathit{conflicts},\Pi,\mu,\mathit{SoC},\Sigma$)\\
        \If {$\mathit{paths} \neq \mathit{UNSAT}$}{
        	\Return $\mathit{paths}$\\
        }
        $\mathit{SoC} \gets \mathit{SoC} + 1$ \\
        $\mu \gets \mu + 1$\\        
    }
}   
 
\SetKwBlock{NRICL}{Heuristic-SMT-CBS-Fixed($\Pi,\mathit{conflicts},\mu,\mathit{SoC},\Sigma$)}{end} \NRICL{
	    \While {$\mathit{TRUE}$}{
	        $\mathcal{F}(\mathit{SoC}) \gets$ encode-Incomplete$(\Pi, \mathit{conflicts},\mu,\mathit{SoC},\Sigma)$\\	    
	        $\mathit{assignment} \gets$ consult-SAT-Solver$(\mathcal{F}(\mathit{SoC}))$\\
	        \If {$\mathit{assignment} \neq \mathit{UNSAT}$}{
	            $\mathit{paths} \gets$ extract-Solution$(\mathit{assignment})$\\
	            $\mathit{collisions} \gets$ validate($\mathit{paths}$)\\
                   \If {$\mathit{collisions} = \emptyset$}{
                      \Return $(\mathit{paths},\mathit{conflicts})$\\
                   }
                   \For{each $(a_i,a_j,v,t) \in \mathit{collisions}$}{
                      $\mathcal{F}(\mathit{SoC}) \gets \mathcal{F}(\mathit{SoC}) \cup \{\neg \mathcal{X}_v^t(a_i) \vee \neg \mathcal{X}_v^t(a_j)$\}\\
                      $\mathit{conflicts} \gets \mathit{conflicts} \cup \{[(a_i,v,t),(a_j,v,t)]\}$
                   }
                   \For{each $a_i \in R$}{
                      $\pi(a_i) \gets$ new-AND-Path($\Pi(a_i), \mathit{conflicts}(a_i),\mu$) \\
                      \If{$\pi(a_i) \neq []$}{
                        $\Pi(a_i) \gets \Pi(a_i) \cup \{\pi(a_i)\}$
                      }
                      \Else{
                        $\Pi(a_i) \gets \{$all paths as in MDD$\}$
                      }
                   }
               }
               \Else{
                   \Return {($\mathit{UNSAT}, \mathit{conflicts}$)}\\
               }
          }
}
\caption{SMT-based optimal MAPF solver with heuristically guided sparsification of the set of candidate paths.} \label{alg-heuristic-SMTCBS}
\end{footnotesize}
\end{algorithm}

The sparse set of candidate paths is used to build an incomplete Boolean MAPF model $\mathcal{F}(\mathit{SoC})$ (line 14) which is subsequently solved by the SAT solver. If a satisfying truth-value assignment of $\mathcal{F}(\mathit{SoC})$ is found (line 16), then it has to be interpreted and checked against MAPF rules, that is, it is checked for collisions between agents (line 18). If there are no collisions, the algorithm can return a valid MAPF solution (line 20). If a collision is detected (lines 21-28), a proper treatment must be applied.

This includes {\bf (1)} extending the model $\mathcal{F}(\mathit{SoC})$ with a collision elimination constraint, a disjunction forbidding a pair of conflicting actions to take place at once (lines 21-23) and {\bf (2)} and extending the sparse set of candidate paths $\Pi$ with a new path reflecting newly discovered conflicts (line 24-29). It may however happen that no new path could be found after which the algorithm switches to SMT-CBS mode using the full MDD as the set of candidate paths.

Without proof let us state the following proposition summarizing the soundness and optimality of the Heuristic-SMT-CBS algorithm. Let us note that unsolvable input MAPF can be detected in advance by some polynomial-algorithm such as Push-and-Rotate \cite{DBLP:journals/jair/WildeMW14}.
\begin{proposition}
Heuristic-SMT-CBS finds sum-of-costs optimal solution for a solvable input MAPF $\Sigma$. $\blacksquare$
\end{proposition}

Path included into $\Pi$ suggested by the AND-path selection can be found by various shortest path algorithms like A* that take into account the set of conflicts.





\section{Experimental Evaluation}

We performed an experimental evaluation of Heuristic-SMT-CBS and compared it against SMT-CBS, and Sparse-SMT-CBS on the standard benchmarks from \texttt{movingai.com} \cite{DBLP:conf/ijcai/BoyarskiFSSTBS15,DBLP:journals/ai/SharonSGF13,DBLP:journals/tciaig/Sturtevant12}. To provide broader perspective outside compilation-based techniques we also include a comparison with a vanilla implementaiton of the CBS algorithm, a search-based algorithm. A representative part of the results is presented in this section.

\subsection{Benchmarks and Setup}

We implemented Heuristic-SMT-CBS in C++ using the existing implementations of Sparse-SMT-CBS and SMT-CBS. The existing implementations of Sparse-SMT-CBS and SMT-CBS were used in the evaluation as well. All variants of SMT-CBS algorithms are built on top of the Glucose 3.0 SAT solver \cite{DBLP:journals/ijait/AudemardS18} that still ranks among the best SAT solvers according to recent SAT solver competitions \cite{DBLP:conf/aaai/BalyoHJ17}. The SAT solver is directly integrated with the main MAPF solver through its API (no call of the SAT solver as an external program occurs).

As for CBS, we used implementation from \cite{DBLP:conf/socs/BarrerSSF14} written in C\#. Although this implementation of CBS does not use recent MAPF heuristics like rectangle reasoning \cite{DBLP:conf/ijcai/LiFB0K19} we consider it suitable for our experiments as implementations of both Sparse-SMT-CBS and SMT-CBS also do not rely on heuristics derived from the knowledge that the problem instance takes place on a grid.

The Glusoces SAT solver we used supports incremental mode which means that when the formula is extended with new variables and clauses the solver does not need to start from scratch but it can use the learned clauses from its previous runs. This incremental mode is highly utilized across the low-level phase of SMT-CBS, Sparse-SMT-CBS, as well as Heuristic-SMT-CBS whenever the target Boolean formula is extended with new conflict elimination clauses. However, each new iteration of the main loop starts with new instance of the SAT solver since establishing an alleviated cost bound cannot be done without significant changes in the formula which can be done more easily for a completely fresh formula \footnote{Incremental mode that works across iterations of the main loop can be implemented by using assumptions through which clauses can be deactivated which effectively allows for removal of clauses from the formula.}.

All experiments were run on system consisting of Xeon 2.8 GHz cores, 32 GB RAM, running Ubuntu Linux 18.


The experimental evaluation has been done on diverse instances consisting of 4-connected {\em grid} maps ranging in sizes from small to relatively large. The set of instances is identical to those used for testing Sparse-SMT-CBS in \cite{DBLP:conf/iros/Surynek21}.

Grid maps are used for discretization of real environments. Agents traverse maps using orthogonal movements (diagonal moves are not used) hence unit time per move can be realistically assumed. Moreover this discretization turned out to be sufficient for execution of plans with real agents as shown in our related research on simulations with small mobile robots \cite{DBLP:conf/smc/ChudyPS20}.

\begin{figure}[t]
    \centering
    \includegraphics[trim={2.5cm 14.5cm 2cm 2.5cm},clip,width=0.95\textwidth]{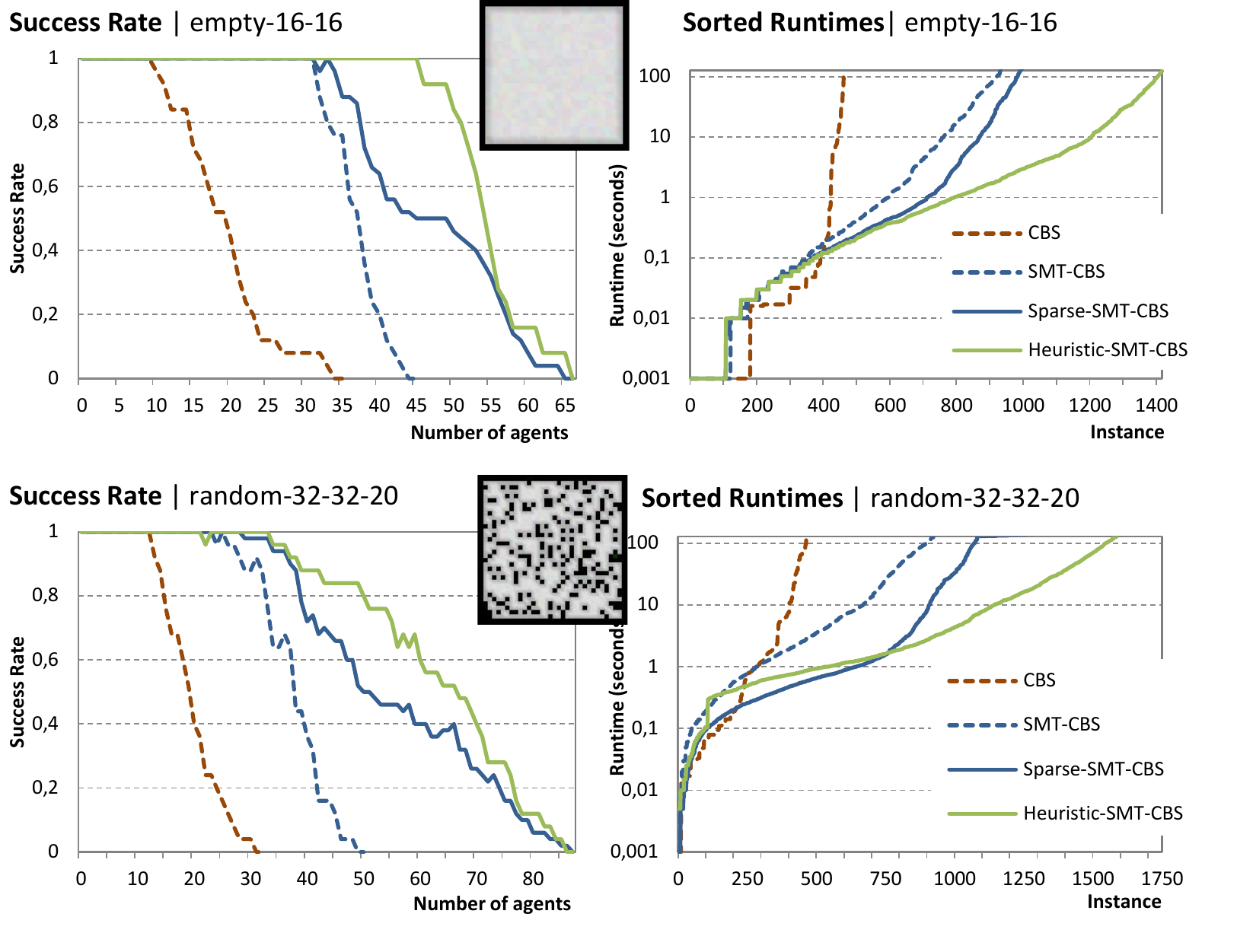}
    \vspace{-0.5cm}
    \caption{Success rate and runtime comparison on small-sized maps.}
    \label{expr-small}
\end{figure}

We varied the number of agents to obtain instances of various difficulties while initial and goal configurations of agents were generated according to scenarios provided on \texttt{movingai.com}. Depending on the map size, we generated instances with up to 64 (small maps) or 128 (large maps) agents and 25 different instances per number of agents. The time limit in all test was set to 128 seconds. Presented results were obtained from instances solved within this timeout.

\subsection{Runtime Results}

\begin{figure}[h]
    \centering
    \includegraphics[trim={2.5cm 14.5cm 2cm 2.5cm},clip,width=0.95\textwidth]{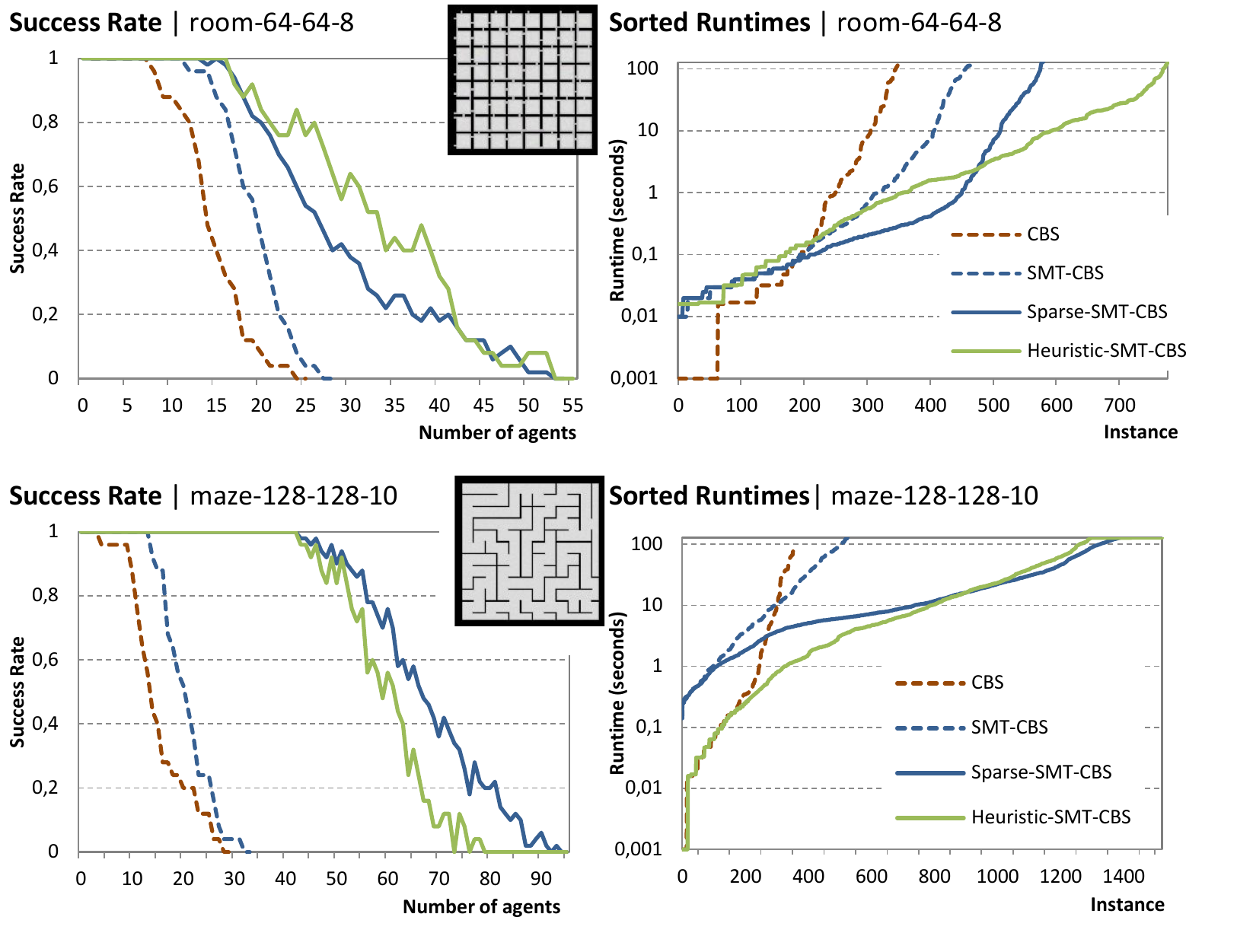}
    \vspace{-0.5cm}
    \caption{Success rate and runtime comparison on medium-sized maps.}
    \label{expr-medium}
\end{figure}

Results are presented in Figures \ref{expr-small}, \ref{expr-medium}, and \ref{expr-large}. We present {\em success rate} and {\em sorted runtimes}. Success rate shows the ratio of instances solved under the time limit of 128 seconds out of 25 instances per number of agents. Sorted runtimes are inspired by the cactus plots from the SAT Competition \cite{DBLP:conf/aaai/BalyoHJ17}.

The cactus plot for runtime has been generated by taking runtimes of all instances solved under the time limit by the given algorithm and sorting them along x-axis; so the x-th data-point represents the runtime of x-th fastest solved instance by the given algorithm. The faster algorithm hence typically yields to a lower curve in the cactus plot but differences for various regions of the plot can be observed too.

The general trend observable across almost all maps from success rate and sorted runtimes is that Heuristic-SMT-CBS represents an improvement over both SMT-CBS and Sparse-SMT-CBS. The one exception is the \texttt{maze-128-128-10} map where Heuristic-SMT-CBS lost to Sparse-SMT-CBS but the gap is relatively small and SMT-CBS performs significantly worse than other two variants of SMT-CBS. The behavior of Heuristic-SMT-CBS on the \texttt{maze-128-128-10} map can be explained by frequent non-existence of path generated by the AND-path selection heuristic.

Sorted runtimes comparison also shows that in Heuristic-SMT-CBS dominates in instances of easy and medium difficulty. However as the difficulty of instances grows often Heuristic-SMT-CBS become worse than Sparse-SMT-CBS. This behavior can be explained by significant reduction of the size of SMDD when the number of collisions is not high. In such a case Heuristic-SMT-CBS constructs only small SMDD which decreases overall solving runtime. On the other hand when there are many collisions between agents, the algorithm switches to full MDD, that is to the identical mode as SMT-CBS and all previous growing of SMDD becomes an overhead.

Comparison with CBS shows an expected trend that due to its negligible overhead, the CBS algorithm dominates in easy instances where SMT-CBS, Sparse-SMT-CBS, and Heuristic-SMT-CBS need to deal with complex instance compilation process that eventually does not pay off. As difficulty of instances grow, the advanced learning mechanisms and intelligent space search pruning implemented in SAT solvers has greater impact resulting in better performance of compilation-based solvers.

A significant shortcoming  of the SMT-CBS solver is that in large grid maps like {\texttt{warehouse-10-20-10-2-1} or {\texttt{lak303d}, the solver generates formulae that are too large which causes degradation in performance. According to our experiments it seems that using sparse set of candidate paths in SMT-CBS helps the solver especially on large maps - there is a relatively big gap between SMT-CBS and its variants with sparsification.

The reason is that the size of MDDs for large maps is often prohibitive and causes significant difficulty for the SAT solver as it needs to deal with a large formula derived from MDD. Sparsification enables choosing few promising paths that are represented in SMDDs which improves both the Boolean formula construction process as well as its solving. Yet fewer paths represented in SMDDs generated by the AND-path selection heuristic leads to an improvement over Sparse-SMT-CBS.

\begin{figure}[h]
    \centering
    \includegraphics[trim={2.5cm 14.5cm 2cm 2.5cm},clip,width=0.95\textwidth]{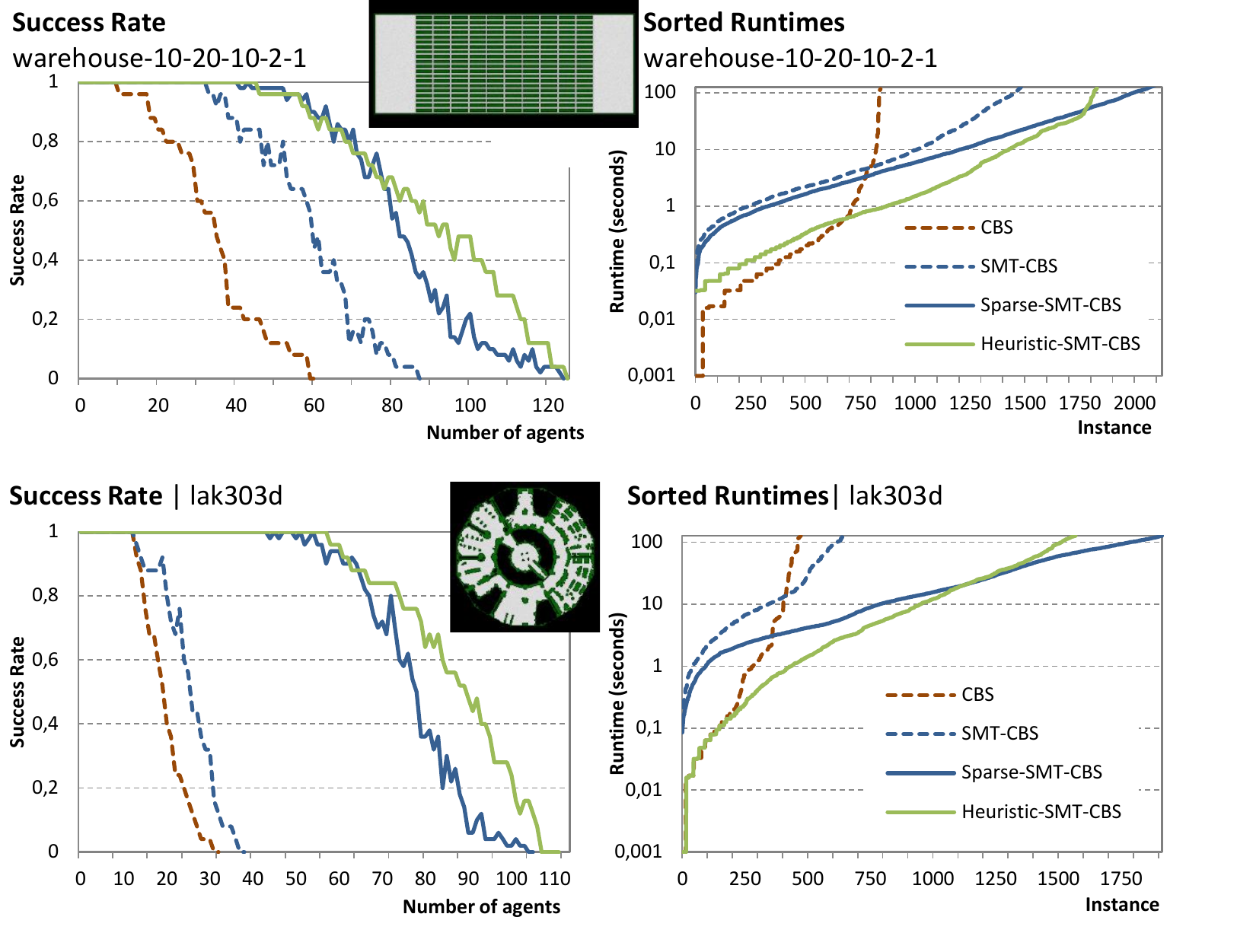}
    \vspace{-0.5cm}
    \caption{Success rate and runtime comparison on large-sized maps.}
    \label{expr-large}
\end{figure}

\section{Conclusion}

We suggested a new heuristically guided selection of candidate paths for agents in sparsification schemes for SAT-based approach to multi-agent path planning (MAPF). The technique aims on currently the most significant drawback of compilation-based approaches to MAPF which is the performance on large graphs.

The previous sparsification technique suggests to construct the target Boolean formula according to a sparse set of candidate paths for each agent which leads to constructing smaller formulae that can be answered by the SAT solver faster. The sparsification technique integrated into SMT-CBS, a state-of-the-art SAT-based optimal algorithm for MAPF, leads to significant performance improvements against the original SMT-CBS.

However, the original sparsification technique still generates large sets of candidate paths. We mitigated this drawback in this work by suggesting a heuristic called {\em AND-path selection}. AND-path selection searches for the most constrained path with respect to discovered collisions between agents. This path when represented multi-value decision diagram (MDD) yields small set of candidate path that still provides enough movement freedom for the agent.

According to our experiments on a number of benchmarks, the new algorithm called Heuristic-SMT-CBS that represents an integration of the AND-path selection into SMT-CBS,  significantly outperform Sparse-SMT-CBS, a variant of SMT-CBS with previous sparsification technique.

For future work we plan to further generalize the concept of sparsification and heuristically guided selection of candidate paths for different variants of MAPF.

\section*{Acknowledgements}
\noindent 
This research has been supported by GA\v{C}R - the Czech Science Foundation, grant registration number 22-31346S.

\bibliographystyle{splncs04}
\bibliography{references}

\end{document}